%% file: main.tex
\documentclass[10pt,twocolumn,letterpaper]{article}

\usepackage{cvpr}              %

\input{preamble}
\definecolor{cvprblue}{rgb}{0.21,0.49,0.74}
\usepackage{pifont}
\newcommand{\cmark}{\ding{51}} %
\newcommand{\xmark}{\ding{55}} %

\usepackage[pagebackref,breaklinks,colorlinks,citecolor=cvprblue]{hyperref}

\usepackage{url}
\usepackage[utf8]{inputenc} %
\usepackage[T1]{fontenc}    %

\usepackage{booktabs}       %
\usepackage{amsfonts}       %
\usepackage{nicefrac}       %
\usepackage{microtype}      %
\usepackage{xcolor}         %
\usepackage{graphicx}
\usepackage{amsmath}
\usepackage{bm}
\usepackage{multirow}
\usepackage{tabu}
\usepackage{siunitx}
\usepackage{adjustbox}
\usepackage{subcaption}
\usepackage{colortbl}
\usepackage{color}
\usepackage{subcaption}
\usepackage{pifont}
\definecolor{Gray}{gray}{0.95}
\definecolor{Cyan}{rgb}{0.88,1,1}
\usepackage{soul} 
\usepackage{framed}
\usepackage{wrapfig}
\usepackage{adjustbox}

\usepackage{tabu}           %
\usepackage{multirow}
\usepackage{booktabs}
\usepackage{makecell}

\def\paperID{5990} %
\def\confName{CVPR}
\def\confYear{2026}

\title{USV: Unified Sparsification for Accelerating Video Diffusion Models}

\author{
Xinjian Wu$^{1,2,*}$ \quad 
Hongmei Wang$^{2}$ \quad
Yuan Zhou$^{2,\dagger}$ \quad
Qinglin Lu$^{2,\S}$ \\
$^1$University of Chinese Academy of Sciences \quad
$^2$Tencent Hunyuan \quad \\
}

\begin{document}
\maketitle
\footnotetext[1]{$*$ Work done during internship at Tencent Hunyuan. \\
\hphantom{$*$\ }\quad \quad Email: \texttt{nuaawxj@gmail.com}}
\footnotetext[2]{$\dagger$ Project leader.}
\footnotetext[3]{$\S$ Corresponding author.}

\input{sec/0_abstract}    
\input{sec/1_intro}

\input{sec/2_related_work}
\input{sec/3_method}
\input{sec/4_experiments}
\input{sec/5_conclusion}

{   
    \small
    \bibliographystyle{ieeenat_fullname}
    \bibliography{main}
}

\end{document}

%% file: sec/0_abstract.tex
\begin{abstract}
The scalability of high-fidelity video diffusion models (VDMs) is constrained by two key sources of redundancy: the quadratic complexity of global spatio-temporal attention and the computational overhead of long iterative denoising trajectories. Existing accelerators---such as sparse attention and step-distilled samplers---typically target a single dimension in isolation and quickly encounter diminishing returns, as the remaining bottlenecks become dominant. In this work, we introduce \textbf{USV} (Unified Sparsification for Video diffusion models), an end-to-end trainable framework that overcomes this limitation by jointly orchestrating sparsification across both the model's internal computation and its sampling process. USV learns a dynamic, data- and timestep-dependent sparsification policy that prunes redundant attention connections, adaptively merges semantically similar tokens, and reduces denoising steps, treating them not as independent tricks but as coordinated actions within a single optimization objective. This multi-dimensional co-design enables strong mutual reinforcement among previously disjoint acceleration strategies. Extensive experiments on large-scale video generation benchmarks demonstrate that USV achieves up to \textbf{83.3$\times$} speedup in the denoising process and \textbf{22.7$\times$} end-to-end acceleration, while maintaining high visual fidelity. Our results highlight unified, dynamic sparsification as a practical path toward efficient, high-quality video generation.
\end{abstract}

%% file: sec/1_intro.tex
\section{Introduction}
\label{sec:intro}

The advent of video diffusion models (VDMs)~\cite{peebles2023scalable,wan2025wan,kong2024hunyuanvideo} marks a major milestone in generative video synthesis, enabling high-fidelity, temporally coherent video generation from textual descriptions and unlocking applications in filmmaking, education, and virtual or augmented reality. However, scaling VDMs to practical, large-scale deployment remains challenging. The computational and memory costs of these models grow rapidly with video resolution and length, and two core bottlenecks dominate: transformer backbones rely on global spatio-temporal self-attention with quadratic complexity in the number of tokens, and diffusion-based sampling is inherently iterative, requiring dozens to hundreds of denoising steps. As a result, even modest resolutions and durations can require minutes of inference on high-end GPUs, severely limiting interactive or real-time use cases.

\begin{figure}[t]
    \centering
    \includegraphics[width=\linewidth]{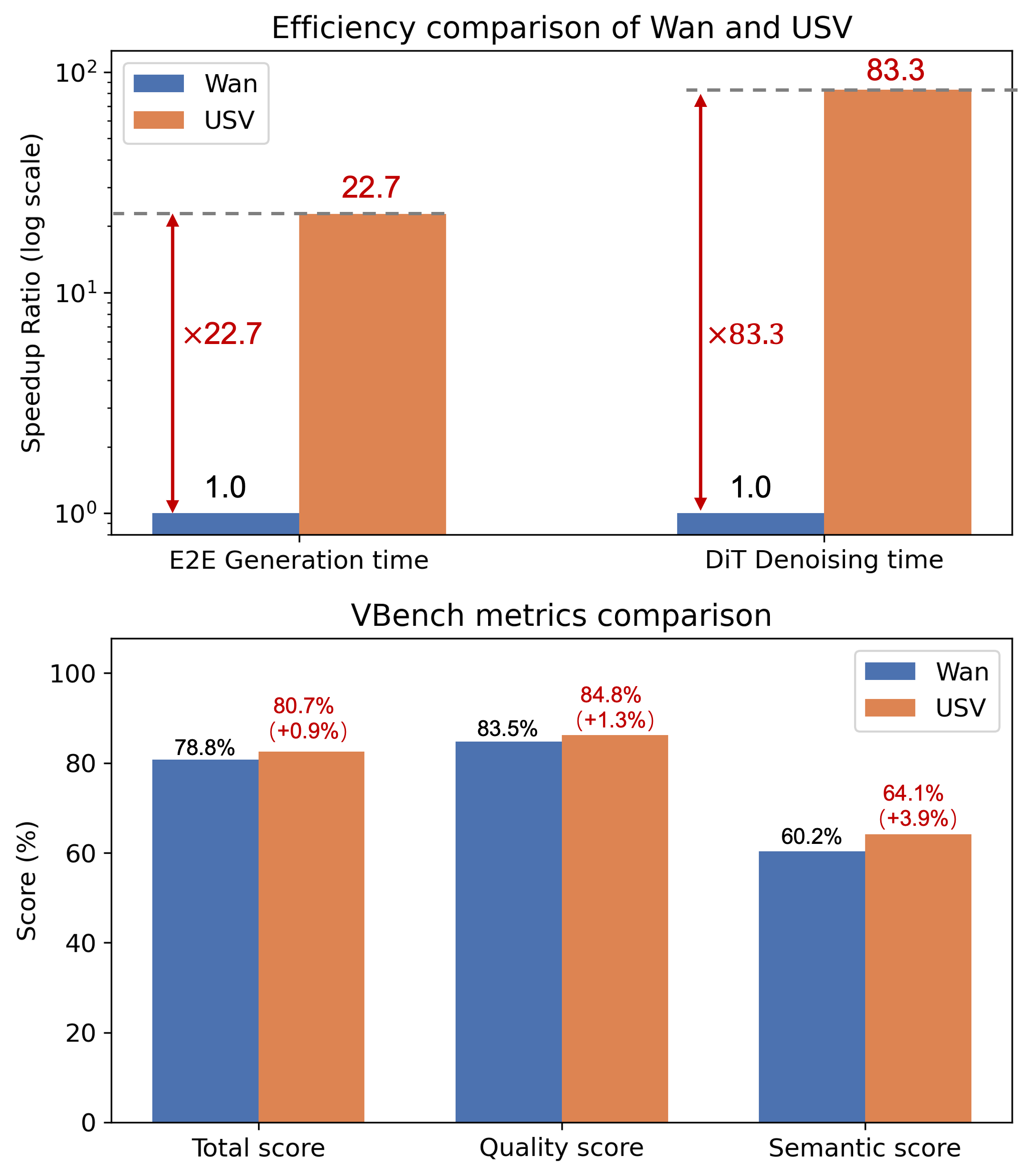}
    \caption{
    (a) Speedup of USV over Wan2.1-1.3B at 480p. Log-scale ratios for end-to-end (E2E) generation 
    and DiT denoising time, normalized to Wan as $1\times$. USV achieves over $20\times$ end-to-end 
    and $80\times$ denoising speedup.
    (b) VBench comparison showing that USV maintains or slightly improves total, quality, 
    and semantic scores compared to the original model.
    }
    \label{fig:intro_eff_vbench}
\end{figure}

To address these challenges, the community has mainly pursued two distinct acceleration directions. One line of work focuses on \textbf{architectural efficiency}~\cite{zhang2025vsa,zhang2025fast,zhan2025bidirectional,wangpt,wu2025vmoba}, introducing sparse attention mechanisms---such as fixed patterns, low-rank approximations, or learned sparsity---to reduce the cost of the transformer backbone. Another line targets \textbf{sampling efficiency}~\cite{luo2023latent,liu2023flow,yin2024onestep,yin2024improved,lin2025diffusion}, primarily through knowledge distillation techniques that condense the multi-step denoising process into far fewer steps or re-parameterize the sampling trajectory. While each direction yields substantial gains, they are typically developed and applied in isolation, optimizing only a single dimension of the overall computation and becoming vulnerable to \emph{diminishing returns}: once attention is aggressively sparsified, token redundancy often emerges as the next bottleneck, and distilling a model that still suffers from architectural inefficiencies leaves much acceleration potential untapped. In short, the limitation lies not in the absence of strong accelerators, but in the lack of a holistic strategy that coordinates multiple, orthogonal sources of redundancy.

Recent efforts such as \textit{FastVideo}~\cite{fastvideo2024} and \textit{Blade}~\cite{gu2025video} begin to bridge architectural and sampling efficiency, hinting at the benefits of joint optimization across components. These methods show that integrating acceleration techniques can yield significant gains in speed and stability. However, their designs remain \textbf{limited in scope}---typically coupling only two aspects (e.g., attention sparsity with step distillation) and applying them through \textbf{fixed or decoupled heuristics}. In particular, token-level redundancy and the interaction between internal architectural sparsity and external denoising dynamics are still largely underexplored. This motivates a more \textbf{comprehensive and unified approach} that systematically coordinates multiple dimensions of sparsification within a single, trainable framework.

In this work, we argue that breaking this performance ceiling requires the \textbf{synergistic co-design} of acceleration techniques and, crucially, a learned \emph{dynamic} policy to coordinate them, rather than applying them in isolation. We present \textbf{USV} (Unified Sparsification for Video diffusion models), a unified sparsification framework that orchestrates optimization across both the model's internal computational pathways and its external sampling trajectory. Unlike methods that apply sparsity post-hoc or in a fixed, heuristic manner, USV is \textbf{end-to-end trainable} and learns a \emph{dynamic, data- and timestep-dependent sparsification policy}. Concretely, USV prunes redundant attention connections, adaptively merges semantically similar tokens, and leverages distillation to shorten the denoising chain---not as separate tricks, but as complementary actions within a single synergistic optimization objective. These dynamic sparsification decisions reinforce one another: compact token representations learned via merging further stabilize and enhance distillation, while the shortened sampling trajectory, in turn, reshapes the attention patterns the model needs to focus on.

As shown in Figure~\ref{fig:intro_eff_vbench}, extensive experiments on large-scale video generation benchmarks validate the efficacy of this unified, dynamic approach. USV achieves a \textbf{83.3$\times$} speedup in the denoising process and a \textbf{22.7$\times$} end-to-end acceleration, while maintaining competitive visual fidelity compared to full-attention baselines. These results show that USV substantially advances the state of the art in efficiency without compromising quality.
Our core contributions are summarized as follows:
\begin{itemize}
    \item We identify and analyze the key limitation of isolated acceleration strategies in VDMs, namely their diminishing returns caused by unaddressed, orthogonal sources of redundancy.
    \item We propose \textbf{USV}, a unified framework that, for the first time, \emph{jointly} sparsifies attention, tokens, and sampling steps in an end-to-end trainable manner for video diffusion models, governed by a learned dynamic sparsification policy.
    \item We demonstrate through rigorous experimentation that USV achieves very high acceleration rates while preserving visual fidelity, establishing multi-dimensional co-design with dynamic sparsification as a practical and effective path toward efficient, high-quality video generation. 
\end{itemize}

\begin{figure*}[t]
    \centering
    \includegraphics[width=\textwidth]{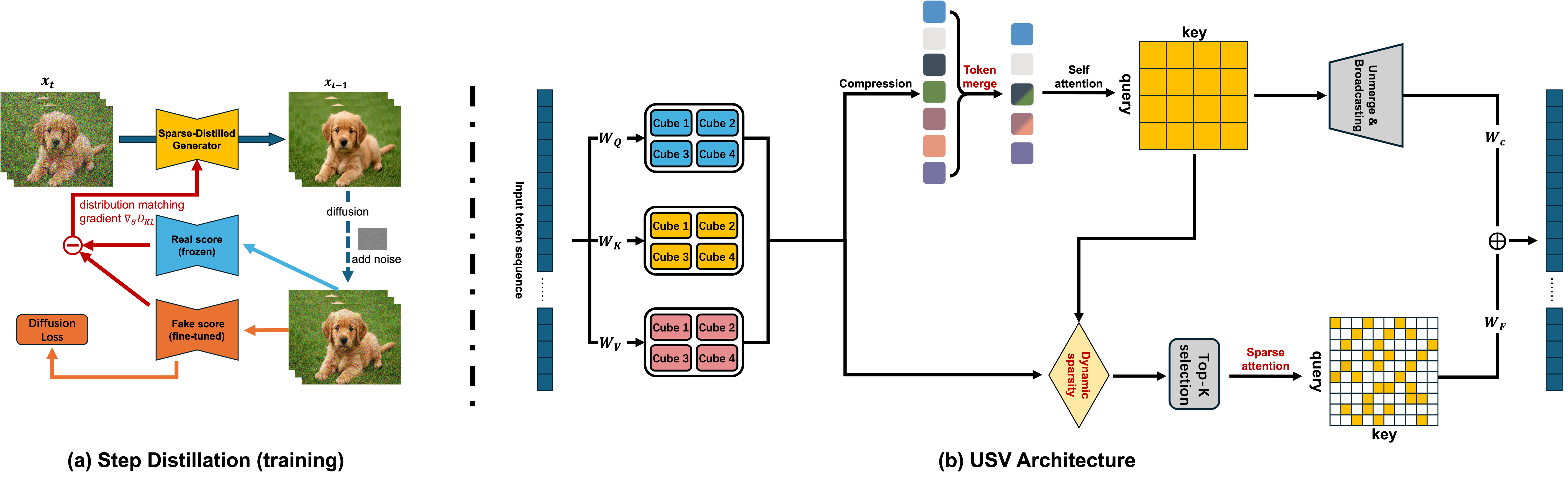}
    \caption{
    Overview of USV. \textbf{Left:} unified sparse distillation.
    A sparse-distilled generator is trained to match a full-attention teacher
    via a distribution-matching gradient from a frozen real score network,
    while a fake score network is fine-tuned with a diffusion loss.
    \textbf{Right:} per-block dynamic sparsification.
    Queries, keys, and values are partitioned into local 3D cubes and passed
    through a token merging module to aggregate redundant tokens.
    An entropy-aware dynamic sparsity scheduler then allocates sparsity between
    local cube attention and long-range sparse attention (via top-$k$ cube
    selection). The outputs are unmerged, broadcast back to the full token
    grid, and fused to produce the final block output.
    }
    \label{fig:usv_overview}
\end{figure*}

%% file: sec/2_related_work.tex
\section{Related Work}
\label{sec:related work}

\subsection{Video Diffusion Models}

Diffusion-based video generation has rapidly advanced from early pixel-space models to large-scale text-to-video systems built on diffusion transformers (DiTs). Early works extend image diffusion models to the temporal domain by inflating 2D UNets or DiTs with temporal layers and finetuning on video data, enabling text-conditioned video synthesis with increasing resolution and duration~\cite{peebles2023scalable,lilian2024videodiffusion}. Subsequent models adopt latent video diffusion, operating in a compressed spatio-temporal representation to reduce computation while preserving visual quality~\cite{stable2024svd,ali2023text2video}. Recent large-scale text-to-video systems further push scale and capability by combining powerful vision-language encoders, high-capacity DiTs, and carefully curated video datasets~\cite{cogvideox2024,wan2025wan,kong2024hunyuanvideo}, achieving long, coherent, and high-fidelity videos but at substantial training and inference cost. Our work is complementary to these modeling advances: rather than proposing a new generative architecture, USV targets the scalability bottlenecks of existing VDMs by learning a unified sparsification policy that reduces attention, token, and sampling redundancy.

\subsection{Video-specific Sparse Attention}

Sparse attention has emerged as a key tool for scaling transformers to long spatio-temporal sequences. Classical sparse patterns (e.g., local or blockwise attention) reduce complexity via hand-crafted connectivity, but may underutilize hardware or degrade quality on long-range motion. Recent video-specific sparse attention methods instead learn or design patterns tailored to video DiTs. VSA~\cite{zhang2025vsa} introduces a hierarchical coarse-to-fine scheme that first identifies critical spatio-temporal regions and then applies dense attention only within selected tiles, providing a single trainable kernel that replaces full attention during both training and inference. Complementary work explores bidirectional sparse attention and dynamic key–value pruning for faster video DiT training~\cite{zhan2025bidirectional}, as well as sliding-tile or window-based attention layouts that align with GPU memory hierarchies~\cite{fastvideo2025sta}. These methods demonstrate that video-specific sparse attention can significantly reduce attention FLOPs and wall-clock latency. However, they primarily focus on architectural sparsity inside the transformer backbone: token redundancy and sampling inefficiency remain unaddressed, and sparsity decisions are often decoupled from the dynamics of the denoising process. In contrast, USV treats attention sparsification as one component of a broader, learned sparsification policy that also governs token merging and step reduction.

\subsection{Acceleration via Step Distillation}

Another line of work accelerates diffusion models by reducing the number of denoising steps via distillation or consistency training. Step distillation compresses a multi-step teacher sampler into a student that runs with far fewer steps, sometimes down to one or a handful, while attempting to preserve the teacher’s output distribution~\cite{luo2023latent,liu2023flow,yin2024onestep,yin2024improved}. For video, recent methods extend these ideas to high-resolution text-to-video generation, combining diffusion distillation with adversarial or perceptual objectives to maintain temporal coherence~\cite{lin2025diffusion,ding2024dollar,lin2024dapt,wang2025videoscene,rcm2025video}. These approaches substantially reduce sampling latency but keep the underlying architecture and attention computation largely intact, leaving significant redundancy in token representations and spatio-temporal attention patterns. Frameworks such as FastVideo further integrate post-training step distillation with system-level optimizations for video diffusion pipelines~\cite{fastvideo2024}. USV differs from prior work by coupling step reduction with architectural sparsity and token merging under a single, end-to-end trainable, dynamic policy, allowing the model to co-adapt its internal computation and external sampling trajectory rather than optimizing each component in isolation.

%% file: sec/3_method.tex
\section{Method}
\label{sec:method}

\subsection{Overview}
\label{sec:overview}

As shown in Figure~\ref{fig:usv_overview}, we aim to accelerate VDMs by sparsifying both \emph{what} to compute and \emph{how long} to compute in a unified framework. Concretely, we consider three structural dimensions of sparsity:
(i) \emph{attention sparsity}, i.e., which token pairs are allowed to attend to each other;
(ii) \emph{token sparsity}, i.e., how many spatio-temporal tokens are processed at each layer; and
(iii) \emph{sampling sparsity}, i.e., how many denoising steps are taken along the sampling trajectory.

Let $\mathbf{x}_t$ denote the noisy latent at diffusion step $t$, and let $\mathbf{Z}_t \in \mathbb{R}^{N\times d}$ be its spatio-temporal token representation inside the backbone, where $N = T\times H \times W$ depends on spatial resolution and video length. A dense VDM applies full self-attention over all $N^2$ token pairs inside each transformer block and repeats this for $T$ steps, leading to a computational cost
\begin{equation}
    \mathcal{C}_{\text{dense}}
    \approx \sum_{t=1}^{T} \sum_{l=1}^{L}
    \mathcal{O}\big(N^2 d\big),
\end{equation}
where $L$ is the number of transformer layers.

USV starts from a strong post-training acceleration baseline in the spirit of FastVideo~\cite{fastvideo2024}, which already sparsifies VDMs along two structural dimensions by combining a video sparse attention kernel with a distilled short-step sampler. Building on this sparse-distilled generator, USV introduces two additional components:
\emph{(i) a video token merging module} that explicitly reduces token redundancy across space and time, and
\emph{(ii) an entropy-aware \textbf{dynamic sparsification policy}} that adapts sparsity \emph{per layer, per timestep, and per input} by inspecting the current attention behavior.

Formally, we replace each dense denoising step of the teacher with a sparse student step
\begin{equation}
    \hat{\mathbf{x}}_{t-1}
    = f_\theta\bigl(\mathbf{x}_t, t, c;\, \mathbf{m}^{\text{attn}}_t, \mathbf{m}^{\text{token}}_t\bigr),
\end{equation}
executed under a reduced step schedule $\mathcal{T}'$ obtained via distillation. The control signals
$\mathbf{m}^{\text{attn}}_t$ (for attention sparsity) and $\mathbf{m}^{\text{token}}_t$ (for token sparsity), as well as the effective sampling schedule, are not fixed but are generated by a lightweight policy network
\begin{equation}
    \bigl(\mathbf{m}^{\text{attn}}_t,\, \mathbf{m}^{\text{token}}_t,\, \mathbf{m}^{\text{step}}_t\bigr)
    = \pi_\phi\bigl(\mathbf{x}_t, \mathbf{Z}_t, t, c\bigr).
\end{equation}
At the core of $\pi_\phi$ is an \emph{entropy-based sparsity scheduler} that measures the information entropy of layer-wise attention maps and uses it to decide \emph{how aggressively} to sparsify each layer at each timestep. In other words, attention, token, and step sparsity span the structural dimensions of USV, while the dynamic policy acts as a learned controller that allocates sparsity across layers, timesteps, and inputs in a content- and budget-aware manner.

\subsection{Preliminaries: FastVideo framework}
\label{sec:prelim}

We build USV on top of FastVideo~\cite{fastvideo2024}, which provides
(i) a trainable video sparse attention kernel (VSA)~\cite{zhang2025vsa} and
(ii) a few-step sparse-distilled generator based on DMD2~\cite{yin2024improved}.

\paragraph{Video Sparse Attention (VSA).}
For a transformer layer with input tokens
$\mathbf{Z}_t \in \mathbb{R}^{N \times d}$ at step $t$ and projections
\(
\mathbf{Q} = \mathbf{Z}_t W_Q,\,
\mathbf{K} = \mathbf{Z}_t W_K,\,
\mathbf{V} = \mathbf{Z}_t W_V
\),
dense self-attention computes
\begin{equation}
    \mathrm{Attn}(\mathbf{Z}_t)
    =
    \mathrm{softmax}\!\Big(
        \frac{\mathbf{Q}\mathbf{K}^\top}{\sqrt{d_k}}
    \Big)\mathbf{V},
\end{equation}
with complexity $\mathcal{O}(N^2 d)$.  
VSA introduces a learned binary mask
$\mathbf{M}^{(l)} \in \{0,1\}^{N \times N}$ for layer $l$ and uses
\begin{equation}
    \mathrm{VSA}(\mathbf{Z}_t)
    =
    \mathrm{softmax}\!\Big(
        \frac{\mathbf{Q}\mathbf{K}^\top
              + \log \mathbf{M}^{(l)}}{\sqrt{d_k}}
    \Big)\mathbf{V},
\end{equation}
where $\mathbf{M}^{(l)}_{ij}=0$ masks disallowed query-key pairs.
In the FastVideo student, all self-attention blocks are replaced by VSA with a
\emph{fixed} sparsity pattern $\mathbf{M}^{(l)}$ at inference time.

\paragraph{Sparse distillation with DMD2.}
Let $\epsilon_{\text{tea}}$ be a long-step teacher denoiser with full attention
and $T$ diffusion steps, and $\epsilon_{\theta}$ a few-step student denoiser
that uses VSA and runs on a reduced step set
$\mathcal{T}' \subset \{1,\dots,T\}$.
FastVideo trains $\epsilon_{\theta}$ using DMD2-style sparse distillation with
a frozen \emph{real} score network $s_{\text{real}}$ and a trainable
\emph{fake} score network $s_{\text{fake}}$, both with dense attention.
Given a noisy sample $\mathbf{x}_t$ at $t\in\mathcal{T}'$, the student produces
a sparse update, which is re-noised and fed into $s_{\text{real}}$ and
$s_{\text{fake}}$ to obtain a distribution-matching gradient for
$\epsilon_{\theta}$, while $s_{\text{fake}}$ is optimized with a diffusion
loss~\cite{yin2024improved}.  
We denote the resulting few-step, VSA-based student as a sparse-distilled
generator
\(
f_{\text{FV}}(\mathbf{x}_t, t, c)
\)
with fixed step schedule $\mathcal{T}'$ and fixed masks
$\{\mathbf{M}^{(l)}\}_{l=1}^{L}$.

From the perspective of USV, FastVideo specifies a two-dimensional
sparsification of VDMs:
architectural sparsity via VSA (fixed $\mathbf{M}^{(l)}$) and sampling sparsity
via the reduced step set $\mathcal{T}'$.
In the following, we keep $f_{\text{FV}}$ backbone as our starting point and introduce
token-level sparsification and a dynamic, entropy-aware policy $\pi_\phi$ that
replaces these fixed choices with input-, layer-, and timestep-dependent
sparsity controls.

\subsection{Token merging for video}
\label{sec:token}

Even with sparse attention and a short-step sampler, the backbone still processes a fixed number of spatio-temporal tokens $N = T \times H \times W$ at every layer and timestep. Long videos, however, exhibit strong redundancy: neighboring patches within and across frames often carry very similar information. We therefore introduce a \emph{token merging} module that reduces the number of tokens seen by attention, while preserving a dense output via an exact unmerge operator.

\paragraph{3D bipartite partition.}
Let $\mathbf{Z}\in\mathbb{R}^{B\times N\times d}$ denote token features at a given layer and diffusion step, where index $i\in\{1,\dots,N\}$ corresponds to $(\tau_i,x_i,y_i)$ in the video. We partition the $(T,H,W)$ grid into non-overlapping 3D blocks of size $(s_t,s_h,s_w)$ and, in each block, designate one \emph{destination} token and treat the remaining ones as \emph{source} tokens. This induces
\begin{equation}
    \{1,\dots,N\} = \mathcal{A} \cup \mathcal{B},\qquad
    \mathcal{A} \cap \mathcal{B} = \emptyset,
\end{equation}
where $\mathcal{B}$ contains exactly one destination index per block and $\mathcal{A}$ all remaining source indices.

\paragraph{Similarity metric and greedy selection.}
We derive a merging score from attention keys. Let $\mathbf{k}_{h,i}\in\mathbb{R}^{d}$ denote the key of head $h\in\{1,\dots,H\}$ at token $i$. We form head-averaged descriptors
\begin{equation}
    \mathbf{m}_i = \frac{1}{H} \sum_{h=1}^{H} \mathbf{k}_{h,i},
    \qquad i=1,\dots,N,
\end{equation}
and define cosine similarities between sources and destinations as
\begin{equation}
    s_{ij}
    = \Big\langle
        \frac{\mathbf{m}_i}{\|\mathbf{m}_i\|_2},
        \frac{\mathbf{m}_j}{\|\mathbf{m}_j\|_2}
      \Big\rangle,\qquad
    \quad i\in\mathcal{A},\ j\in\mathcal{B}.
\end{equation}
For each source $i\in\mathcal{A}$ we find its best destination and score
\begin{equation}
    j^\star(i) = \arg\max_{j\in\mathcal{B}} s_{ij},\qquad
    u_i = \max_{j\in\mathcal{B}} s_{ij}.
\end{equation}
We sort $\mathcal{A}$ by $u_i$ and select the $r$ most confident sources for merging:
\begin{equation}
    \mathcal{M} = \text{top-}r\,\{i \in \mathcal{A} \mid u_i\},\qquad
    \mathcal{U} = \mathcal{A} \setminus \mathcal{M},
\end{equation}
where $r$ is controlled by a token sparsity ratio $\rho^{\text{token}} \in [0,1]$,
\begin{equation}
    r = \bigl\lfloor \rho^{\text{token}} \cdot N \bigr\rfloor,\qquad r \le |\mathcal{A}|.
\end{equation}

\paragraph{Merge operator.}
For each destination token $j \in \mathcal{B}$, let
\begin{equation}
    \mathcal{M}(j) = \{ i \in \mathcal{M} \mid j^\star(i) = j \}
\end{equation}
be its merged sources. We update destination features by mean aggregation,
\begin{equation}
    \tilde{\mathbf{z}}_j
    = \frac{\mathbf{z}_j + \sum_{i\in\mathcal{M}(j)} \mathbf{z}_i}
           {1 + |\mathcal{M}(j)|},
    \qquad j \in \mathcal{B},
\end{equation}
while unmerged sources keep their original features,
$\tilde{\mathbf{z}}_i = \mathbf{z}_i$ for $i\in\mathcal{U}$. The merged sequence fed into attention is
\begin{equation}
    \tilde{\mathbf{Z}}
    = \bigl[\, \tilde{\mathbf{z}}_i \mid i\in\mathcal{U} \bigr]
      \,\Vert\,
      \bigl[\, \tilde{\mathbf{z}}_j \mid j\in\mathcal{B} \bigr]
    \in \mathbb{R}^{B\times (N-r)\times d},
\end{equation}
so the attention cost scales with $N-r$ instead of $N$.

\paragraph{Unmerge operator and video awareness.}
During merging we store $\mathcal{U}$, $\mathcal{M}$, $\mathcal{B}$ and $j^\star(\cdot)$, which define an exact inverse that restores a dense grid after attention and MLP. Let $\tilde{\mathbf{Z}}'$ denote the updated merged tokens. The recovered tokens $\hat{\mathbf{z}}_k$ at original positions are
\begin{equation}
    \hat{\mathbf{z}}_k =
    \begin{cases}
        \tilde{\mathbf{z}}'_k, & k \in \mathcal{U}, \\
        \tilde{\mathbf{z}}'_k, & k \in \mathcal{B}, \\
        \tilde{\mathbf{z}}'_{j^\star(k)}, & k \in \mathcal{M},
    \end{cases}
\end{equation}
yielding $\hat{\mathbf{Z}}\in\mathbb{R}^{B\times N\times d}$ aligned with the original spatio-temporal grid. Because sources and destinations are restricted to local 3D blocks, merges naturally follow coherent spatio-temporal “tubelets’’ and avoid collapsing distant, unrelated regions, which helps preserve temporal consistency.

In summary, the token merging module builds a local 3D bipartite graph, greedily merges the $r$ most redundant sources into block-wise anchors, and reduces attention cost while preserving dense per-token predictions. The merge ratio $\rho^{\text{token}}$ is later governed by the dynamic sparsification policy (Sec.~\ref{sec:policy}), allowing more aggressive merging at early diffusion steps or high-resolution layers and milder merging when fine details are needed.

\subsection{Entropy-aware dynamic sparsification policy}
\label{sec:policy}

\begin{figure}[t]
    \centering
    \includegraphics[width=\linewidth]{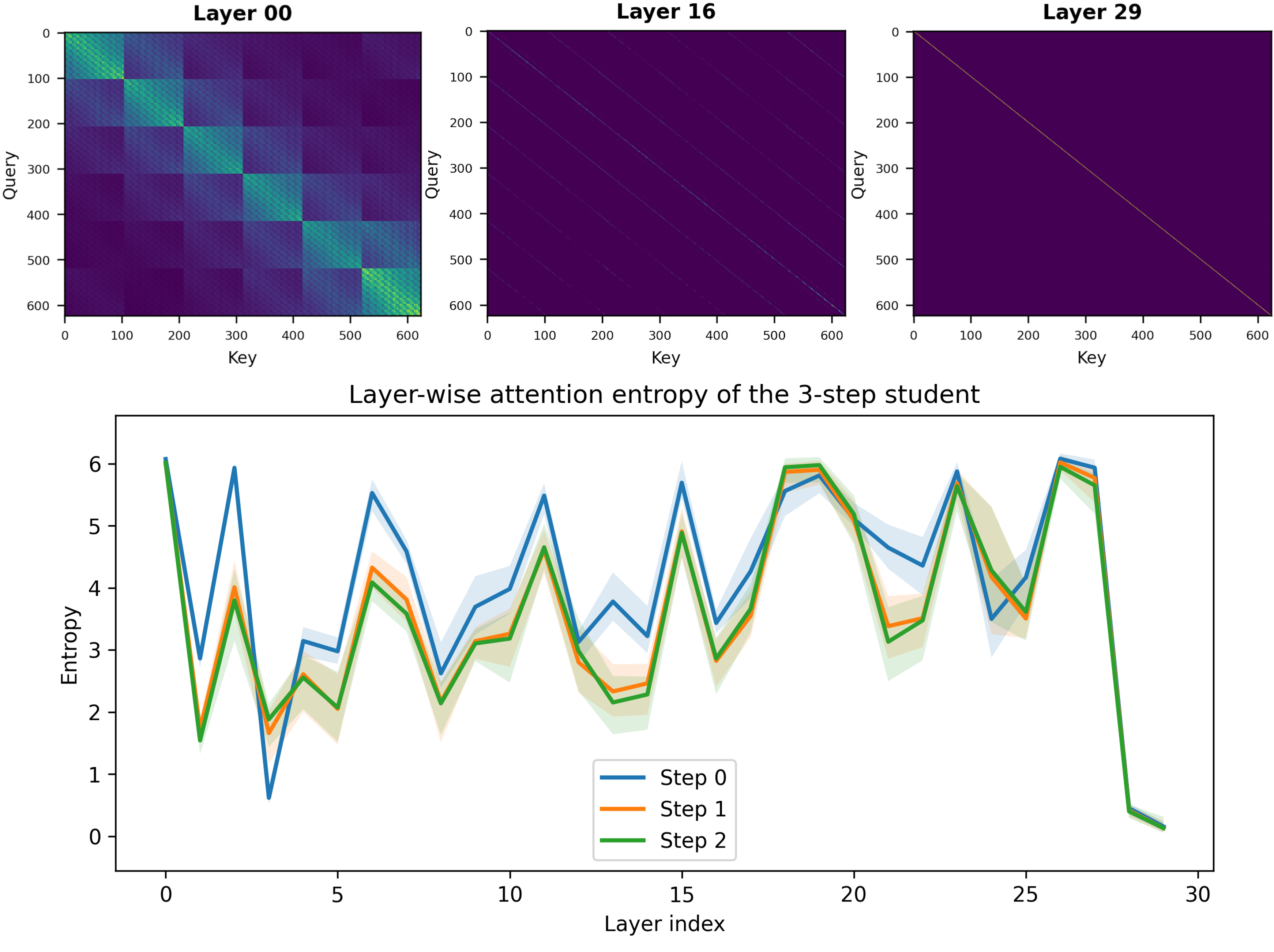}
    \caption{
    Layer-wise attention and entropy of the 3-step sparse-distilled student.
    \textbf{Top:} head-averaged self-attention maps of layers 0, 16, and 29 at a
    fixed diffusion step, showing increasingly concentrated near-diagonal
    patterns in deeper layers. \textbf{Bottom:} layer-wise attention entropy
    over $20$ validation videos; solid curves denote the mean for each step and
    shaded bands the min--max range, highlighting consistently low-entropy,
    highly redundant layers.
    }
    \label{fig:entropy_trends}
\end{figure}

We observe that different layers and timesteps exhibit highly non-uniform
attention entropy (Figure~\ref{fig:entropy_trends}): some layers are consistently
low-entropy (highly concentrated attention, more redundant), while others
remain diffuse. This makes a fixed, one-size-fits-all sparsity schedule
sub-optimal. Instead of using a single global sparsity rate, we design an
\emph{entropy-aware scheduler} that redistributes a pre-defined sparsity
budget across layers and timesteps according to their attention entropy.

\paragraph{Attention entropy.}
For transformer layer $l$ and diffusion step $t$, let
$\mathbf{A}_{t}^{(l)} \in \mathbb{R}^{H \times N \times N}$ denote the
multi-head attention weights (after softmax), where $H$ is the number of
heads. We average over heads and queries to obtain a marginal over keys:
\begin{equation}
    \bar{\mathbf{a}}_{t}^{(l)}(j)
    = \frac{1}{H N}\sum_{h=1}^{H}\sum_{i=1}^{N} \mathbf{A}_{t}^{(l)}(h,i,j),
\end{equation}
and define a normalized entropy
\begin{equation}
    h_{t,l}
    = - \frac{1}{\log N} \sum_{j=1}^{N}
        \bar{\mathbf{a}}_{t}^{(l)}(j)\,\log \bar{\mathbf{a}}_{t}^{(l)}(j)
    \in [0,1].
\end{equation}
Low $h_{t,l}$ means that most queries attend to a small subset of keys
(high redundancy); high $h_{t,l}$ indicates diffuse attention where aggressive
sparsification is risky.

\paragraph{Entropy-aware allocation of sparsity.}
We start from a hand-crafted per-step sparsity schedule
$\{\bar{\rho}^{\text{attn}}_t, \bar{\rho}^{\text{token}}_t\}_{t\in\mathcal{T}'}$
derived from the target compute budget (Sec.~\ref{sec:experiments}). The
entropy-aware scheduler then redistributes this budget across layers. For each
step $t$, we convert entropy into a non-negative importance weight
\begin{equation}
    w_{t,l} = (1 - h_{t,l})^\gamma,
\end{equation}
where $\gamma \ge 1$ controls how strongly low-entropy layers are favored.
We normalize these weights over layers and obtain layer-wise sparsity ratios
\begin{equation}
    \rho^{\text{attn}}_{t,l}
    = \mathrm{clip}\Big(
        \rho^{\text{attn}}_{\min},
        \rho^{\text{attn}}_{\max},\;
        \bar{\rho}^{\text{attn}}_t
        \cdot
        \frac{w_{t,l}}{\frac{1}{L}\sum_{l'=1}^{L} w_{t,l'}}
      \Big),
\end{equation}
and analogously for token sparsity
$\rho^{\text{token}}_{t,l}$, which determines the token budget
$K_{t,l} = \rho^{\text{token}}_{t,l} N$ used by the merging module
(Sec.~\ref{sec:token}). By design, the per-step average of
$\rho^{\text{attn}}_{t,l}$ (and of $\rho^{\text{token}}_{t,l}$) matches the
base schedule $(\bar{\rho}^{\text{attn}}_t,\bar{\rho}^{\text{token}}_t)$,
so the overall compute closely follows the prescribed budget while allocating
stronger sparsity to low-entropy layers and preserving computation where
entropy remains high.

In practice, this entropy-aware scheduler is parameter-free and introduces no
additional learnable modules or loss terms: it simply replaces a fixed sparse
pattern with a content- and timestep-adaptive sparsity allocation driven by
attention entropy.

\subsection{Training Objective and Optimization}
\label{sec:training}

USV trains a sparse student backbone $f_\theta$ and a dynamic policy $\pi_\phi$ within the FastVideo framework. The goal is to preserve the behavior of the teacher model under a target compute budget while allowing the policy to allocate sparsity across attention, tokens, and steps.

\paragraph{Distillation from the FastVideo Teacher.}
The sparse distillation pipeline from FastVideo is used, replacing the student generator with our sparsified model. We maintain a frozen teacher model $f_\varphi$ with full attention, and a trainable student $G_{\theta,\phi}$ that incorporates VSA, token merging, and the entropy-aware policy. The student is updated by matching the trajectory distributions between the teacher and student, while the teacher's frozen model is used to guide the student's optimization.

\paragraph{Budget-Aware Sparsity and Regularization.}
The dynamic policy outputs sparsity ratios for each layer and timestep, which are used to compute a differentiable proxy for the total cost. A budget loss constrains this cost to a user-specified budget, encouraging the policy to adjust sparsity across attention, tokens, and steps. Regularization terms are applied to avoid early collapse and ensure stability across frames.

The overall objective is a combination of distillation, budget loss, entropy regularization, and temporal consistency regularization.

In practice, training proceeds in two stages: first, warming up the student with fixed sparsity, then enabling token sparsity and the policy to gradually optimize for the budget while maintaining the teacher's behavior. During inference, only the sparse student and policy are used.

%% file: sec/4_experiments.tex
\section{Experiments}
\label{sec:experiments}

\subsection{Experimental setup}
\label{sec:setup}

\begin{table*}[t]
    \centering
    \small
    \setlength{\tabcolsep}{5pt}
    \scalebox{1}{
    \begin{tabular}{l|c|c|ccc|cc}
    \hline
    \multirow{2}{*}{Method} 
        & \multirow{2}{*}{Sparsity} 
        & \multirow{2}{*}{Steps} 
        & \multicolumn{3}{c|}{VBench Scores$\uparrow$} 
        & \multicolumn{2}{c}{Speedup $\uparrow$} \\ \cline{4-8}
        & & 
        & Total & Quality & Semantic 
        & E2E & DiT \\ \hline
    Wan2.1 (dense) 
        & 0 
        & 50 
        & 78.8 & 83.5 & 60.2 
        & 1.0$\times$ & 1.0$\times$ \\ \hline
    FastWan (baseline) 
        & 0.80 
        & 3 
        & \textbf{80.8} & 84.5 & \textbf{66.2} 
        & 20.3$\times$ & 73.0$\times$ \\ \hline
    \textbf{USV (ours)} 
        & \textbf{0.95} 
        & 3 
        & {80.7} & \textbf{84.8} & {64.1} 
        & \textbf{22.7$\times$} & \textbf{83.3$\times$} \\ \hline
    \end{tabular}}
    \caption{Comparison with baselines on Wan2.1-1.3B evaluated at 480p resolution with 81-frame sequences (\(\approx\)131K spatio-temporal tokens per sample). USV delivers over \textbf{20$\times$} end-to-end and \textbf{80$\times$} denoising acceleration while maintaining comparable or slightly better VBench Total and Quality scores. Notably, USV sustains strong Semantic alignment despite its significantly higher sparsity,demonstrating the effectiveness of unified sparsification under extreme efficiency.
}
    \label{tab:comparison}
\end{table*}

\begin{figure*}[!h]
    \centering
    \includegraphics[width=\textwidth]{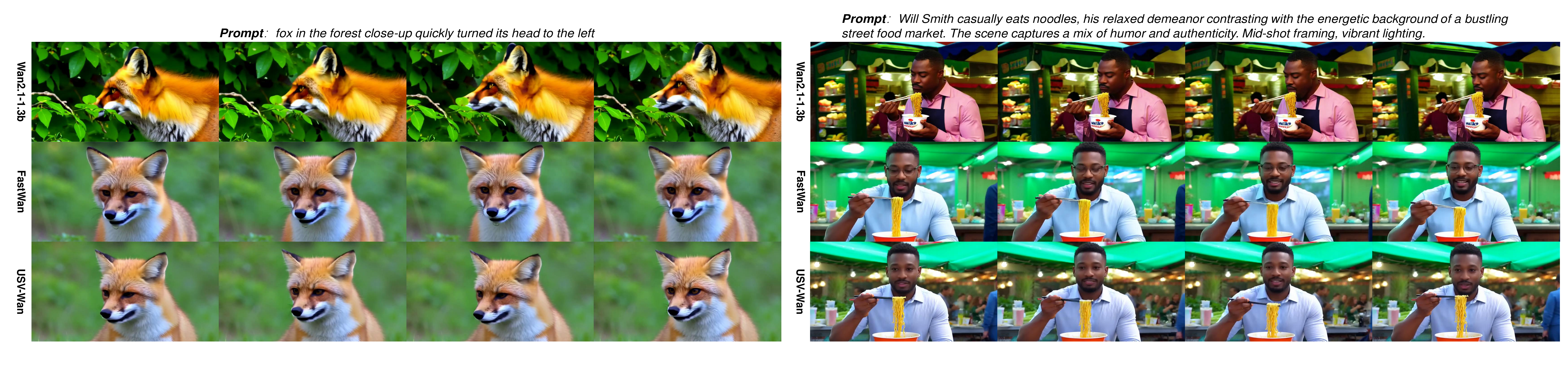}
    \caption{
    \textbf{Qualitative comparison among Wan, FastWan, and our USV.}
    Each row corresponds to a method and each column shows consecutive frames of the same prompt
    (480p, 81 frames; examples from two scenes). 
    The dense Wan baseline produces stable but relatively softer textures. 
    FastWan reduces runtime but exhibits mild temporal flicker and occasional blur due to static sparsity.
    In contrast, \textbf{USV} maintains sharper details and smoother motion across time by 
    dynamically allocating computation along layers and timesteps. 
    The visual trends align with the quantitative results in Table~\ref{tab:comparison}, 
    showing that unified sparsification substantially accelerates video diffusion without compromising perceptual fidelity.
    }
    \label{fig:qualitative}
\end{figure*}

\paragraph{Dataset.}
We conduct all training and evaluation experiments on the Vchitect-T2V-Dataverse dataset~\cite{fan2025vchitect}. This comprehensive benchmark contains a diverse collection of text-video pairs, encompassing a wide variety of scenes, camera motions, and dynamic interactions. For a fair and consistent evaluation, we use the validation split and evaluate all models on an identical set of prompts sampled from the dataset.

\paragraph{Baselines.}
We compare USV against two baselines: (i) the original dense Wan2.1 model with full global attention and the default long-step sampler, and (ii) the FastWan model~\cite{fastvideo2024}, which combines Video Sparse Attention (VSA)~\cite{zhang2025vsa} with a few-step DMD2-based sampler~\cite{yin2024improved}. 

\paragraph{Metrics.}
We evaluate all methods mainly from two aspects: inference efficiency and visual quality. 
For efficiency, we mesure the wall-clock latency on a single NVIDIA H20 GPU. Specifically, we report (i) the end-to-end generation time per video and (ii) the total DiT denoising 
time.  The speedup ratio over the dense Wan baseline is reported for both metrics. 
For visual quality, we adopt VBench-1.0~\cite{huang2024vbench} and report 
its aggregated \emph{Total}, \emph{Quality}, and \emph{Semantic} scores, 
which provide comprehensive metrics across temporal consistency, image quality, 
and text–video alignment.

\paragraph{Implementation details.}
Our implementation is built upon the FastVideo codebase and follows its training recipe unless otherwise specified. We employ the Wan2.1-1.3B backbone and initialize USV from the pre-trained Wan2.1 model. We use AdamW optimizer with cosine learning-rate decay and warm-up, and maintain the same global batch size, number of optimization steps, and total inference FLOPs budget as the FastWan baseline to ensure fair comparison. All experiments are conducted on NVIDIA H20 GPUs.

\subsection{Comparison with baselines}

\paragraph{Quantitative Results.}

Table~\ref{tab:comparison} compares our USV with the dense Wan2.1 model and the accelerated FastWan baseline on 480p, 81-frame videos (\(\sim131\)K tokens per sample). 
USV achieves a remarkable \textbf{22.7$\times$} end-to-end and \textbf{83.3$\times$} denoising acceleration over the full-attention Wan model, 
while maintaining nearly identical VBench quality. 
Compared to FastWan---which already combines sparse attention (VSA) and step distillation (DMD2)---USV further improves efficiency and fidelity through unified sparsification. 
The additional token merging and dynamic sparsity policy yield an extra \textbf{12--15\%} speedup without quality degradation. 
These results confirm that jointly coordinating sparsity across attention, tokens, and sampling steps enables \textbf{super-additive efficiency gains} beyond isolated acceleration strategies.

\paragraph{Qualitative Results.}
Figure~\ref{fig:qualitative} compares Wan, FastWan, and the proposed USV on two prompts, with four consecutive frames shown per method.
The dense Wan baseline is visually stable but produces softer textures.
FastWan, which adopts static sparsity, accelerates inference but introduces mild temporal flicker and occasional blurring.
Benefiting from entropy-aware dynamic allocation, \textbf{USV} preserves sharp details and temporal coherence across frames,
consistent with the speed-quality trade-off in Table~\ref{tab:comparison}.
These results indicate that \emph{unified and dynamic} sparsification is crucial for maintaining perceptual fidelity under extreme efficiency.

\subsection{Ablation studies}
\label{sec:ablations}

\begin{table*}[t]
    \centering
    \small
    \setlength{\tabcolsep}{5pt}
    \scalebox{1.0}{
    \begin{tabular}{l|ccc|ccc|cc}
    \hline
    \multirow{2}{*}{Method} &
    \multicolumn{3}{c|}{Sparsification Dimensions} &
    \multicolumn{3}{c|}{VBench Scores$\uparrow$} &
    \multicolumn{2}{c}{Speedup $\uparrow$} \\ \cline{2-9}
    & Attn & Steps & Tokens &
    Total & Quality & Semantic &
    E2E & DiT \\ \hline
    Wan2.1 (dense) &
    \xmark & \xmark & \xmark &
    78.8 & 83.5 & 60.2 &
    1.0$\times$ & 1.0$\times$ \\

    + VSA only &
    \cmark & \xmark & \xmark &
    79.0 & 83.2 & 60.8 &
    1.7$\times$ & 4.3$\times$ \\

    + DMD only &
    \xmark & \cmark & \xmark &
    79.9 & 84.8 & 60.3 &
    18.2$\times$ & 36.1$\times$ \\

    FastWan (VSA + DMD) &
    \cmark & \cmark & \xmark &
    \textbf{80.8} & 84.5 & \textbf{66.2} &
    20.3$\times$ & 73.0$\times$ \\

    + merge + DMD &
    \xmark & \cmark & \cmark &
    77.1 & 80.8 & 62.2 &
    20.2$\times$ & 42.7$\times$ \\

    \textbf{USV (ours)} &
    \cmark & \cmark & \cmark &
    {80.7} & \textbf{84.8} & {64.1} &
    \textbf{22.7$\times$} & \textbf{83.3$\times$} \\ \hline
    \end{tabular}}
    \caption{
Ablation on sparsification dimensions for Wan2.1-1.3B at 480p. 
We progressively enable sparsity along attention, sampling, and token dimensions. 
Starting from the dense Wan baseline, individual sparsification strategies 
(VSA for attention, DMD for sampling, and token merging for spatio-temporal redundancy reduction) 
each bring partial acceleration and quality variations. 
When jointly applied, our unified USV model (\cmark~in all columns) achieves the best trade-off, 
achieving a \textbf{22.7$\times$} end-to-end and \textbf{83.3$\times$} denoising speedup 
while maintaining high VBench \textit{Total}, \textit{Quality}, and \textit{Semantic} scores.
}

    \label{tab:ablation}
\end{table*}

\paragraph{Effect of multi-dimensional sparsification.}

Table~\ref{tab:ablation} analyzes how different sparsification dimensions contribute to efficiency and generation quality. 
Starting from the dense Wan2.1 baseline (\xmark~in all dimensions), applying sparse attention (VSA) alone moderately improves speed (\(1.7\times\) E2E) and slightly stabilizes visual quality, showing that redundancy in attention maps can be reduced without degradation. 
Step distillation (DMD) alone provides the dominant acceleration (\(18.2\times\) E2E, \(36.1\times\) DiT), confirming that shortening the denoising trajectory is highly effective. 
When both are enabled, FastWan (\cmark~in Attn and Steps) achieves a balanced performance with the best Total and Semantic scores. 
Introducing token merging without attention sparsity (\textit{+merge+DMD}) yields faster inference but suffers noticeable quality loss, suggesting that uncoordinated sparsity may disrupt information flow. 
Finally, our full USV model (\cmark~in all three) combines all dimensions synergistically, delivering the highest overall efficiency (\textbf{22.7$\times$} E2E, \textbf{83.3$\times$} DiT) while maintaining strong perceptual and semantic consistency.

\begin{figure}[t]
    \centering
    \includegraphics[width=\linewidth]{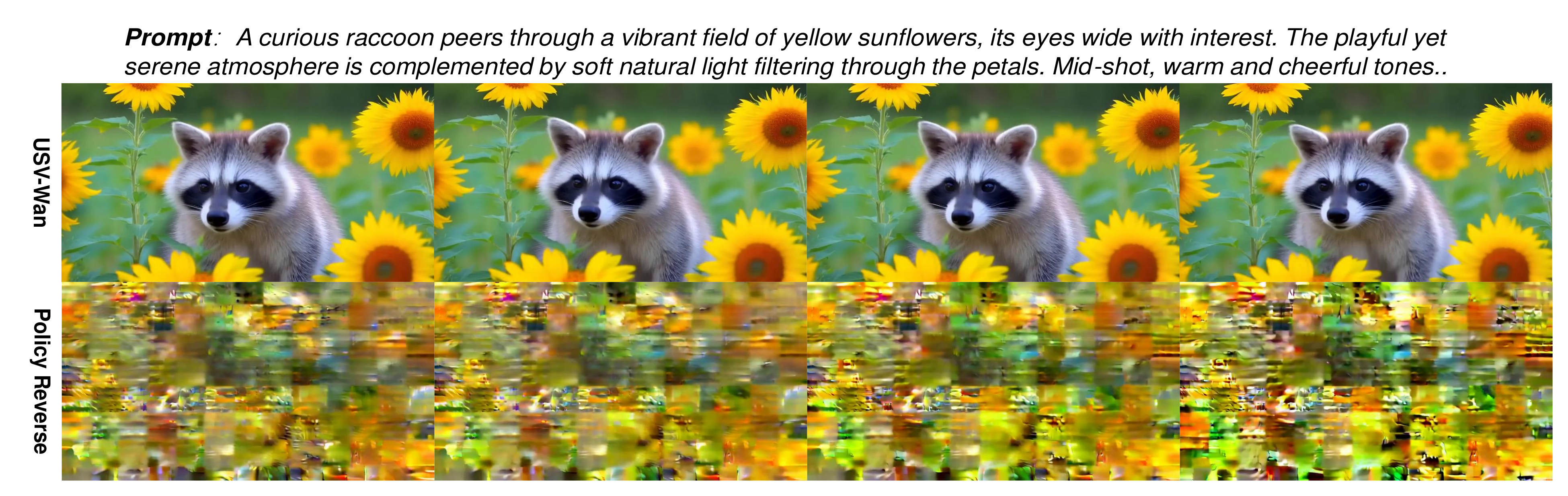}
    \caption{
    \textbf{Effect of dynamic sparsification policy.}
    We compare our learned dynamic sparsification schedule (top) with a reversed static variant (bottom),
    under the same sparsity and compute budget. 
    The reversed policy, which allocates high sparsity to later timesteps, collapses into severe temporal flickering 
    and texture corruption, while our dynamic policy maintains stable structure and fine details across frames.
    This highlights that adaptively allocating sparsity over timesteps is crucial for robust video diffusion.
    }
    \label{fig:dynamic_policy}
\end{figure}

\paragraph{Dynamic sparsification policy.}
To investigate the role of dynamic sparsity allocation, we visualize in Figure~\ref{fig:dynamic_policy} 
the outputs generated by our learned dynamic policy and a reversed static counterpart, 
both trained with identical sparsity ratios and budgets. 
The reversed schedule, which assigns higher sparsity to late denoising steps, 
causes catastrophic degradation and temporal instability, producing inconsistent or fragmented structures. 
In contrast, our entropy-aware dynamic policy adaptively adjusts sparsity per step and per layer, 
preserving texture fidelity and motion continuity throughout the denoising process.
These results demonstrate that static or inverted sparsity fails to generalize across time, 
and that the learned dynamic scheduling is essential for maintaining stability under extreme acceleration.

\begin{figure}[t]
    \centering
    \includegraphics[width=\linewidth]{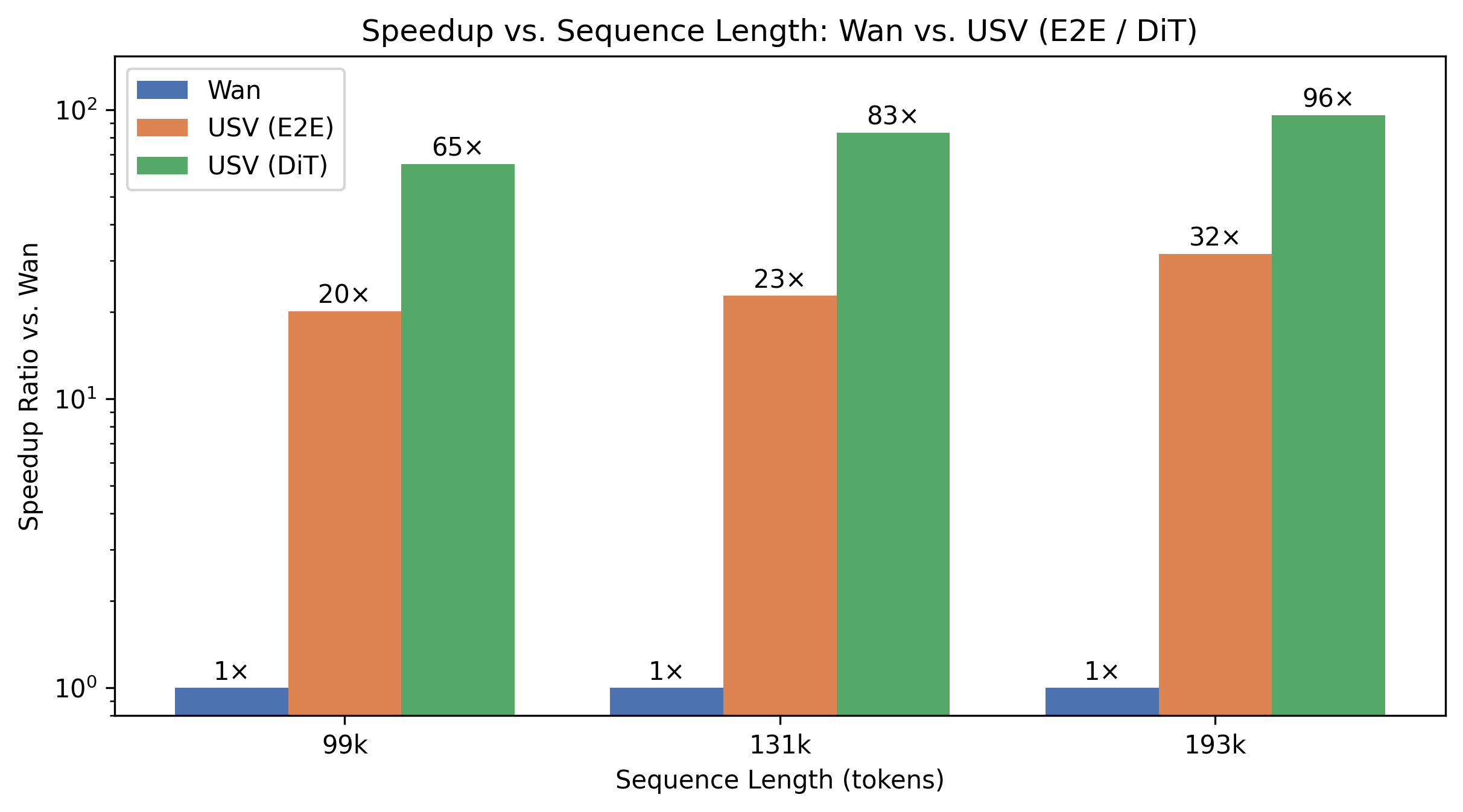}
    \caption{
    \textbf{Speedup vs.\ sequence length on a log-scale axis.}
    For each sequence length (\textbf{99k}, \textbf{131k}, \textbf{193k} tokens), we plot three bars:
    the dense Wan baseline (normalized to \textbf{1$\times$}), and our USV speedups for
    \emph{end-to-end} generation (E2E) and \emph{DiT denoising} (DiT). USV delivers consistent and substantial acceleration, with E2E speedups of \textbf{20$\times$}, \textbf{23$\times$}, and \textbf{32$\times$}. Even more significant gains are observed in the denoising stage, reaching \textbf{65$\times$}, \textbf{83$\times$}, and \textbf{96$\times$} as the sequence length increases. 
    The monotonic increase in speedup with longer sequences highlights the scalability of our sparsification,
    whereas the dense baseline remains fixed at \textbf{1$\times$}.
    }
    \label{fig:speedup_seq}
\end{figure}

\paragraph{Scaling with sequence length.}
Figure~\ref{fig:speedup_seq} reports speedups against the dense Wan baseline across
increasing token sequence lengths (99k, 131k, 193k; log-scale $y$-axis).
Our USV improves both \emph{end-to-end} time (\textbf{20$\times$} $\rightarrow$ \textbf{23$\times$} $\rightarrow$ \textbf{32$\times$})
and \emph{DiT denoising} time (\textbf{65$\times$} $\rightarrow$ \textbf{83$\times$} $\rightarrow$ \textbf{96$\times$})
as the sequence grows, indicating that unified sparsification benefits more from larger spatio-temporal contexts.
By contrast, the dense baseline remains at \textbf{1$\times$}.
These results corroborate our design: attention and token sparsity reduce the dominant quadratic/linear
costs in long sequences, while step sparsity further amplifies the end-to-end gains.

%% file: sec/5_conclusion.tex
\section{Conclusion}
\label{sec:conclusion}
In this paper, we presented USV, a unified framework for multi-dimensional sparsification in video diffusion models. By addressing the key dimensions of attention, token, and sampling sparsity, USV effectively accelerates the denoising process while maintaining high-quality outputs. Our extensive experiments demonstrate the effectiveness of USV in achieving a favorable trade-off between computational efficiency and visual fidelity. Future work will explore further optimizations and potential applications of USV in other generative tasks.